\def\year{2022}\relax
\documentclass[letterpaper]{article} 
\usepackage{aaai21}  
\usepackage{times}  
\usepackage{helvet} 
\usepackage{courier}  
\usepackage[hyphens]{url}  
\usepackage{graphicx} 
\usepackage{natbib}  
\usepackage{caption} 
\usepackage{bm}
\usepackage{amssymb}
\usepackage{latexsym}
\usepackage{dsfont}
\usepackage{multirow}
\usepackage{color}

\title{Memory-Guided Semantic Learning Network for Temporal Sentence Grounding}
\author{Daizong Liu\textsuperscript{\rm 1,2}, Xiaoye Qu\textsuperscript{\rm 2}, Xing Di\textsuperscript{\rm 3}, Yu Cheng\textsuperscript{\rm 4}, Zichuan Xu\textsuperscript{\rm 5}, Pan Zhou\textsuperscript{\rm 1*} \\}
\affiliations{
\textsuperscript{\rm 1}The Hubei Engineering Research Center on Big Data Security, School of
Cyber Science and Engineering, \\ Huazhong University of Science and Technology\\
\textsuperscript{\rm 2}School of Electronic Information and Communication, Huazhong University of Science and Technology\\
\textsuperscript{\rm 3}ProtagoLabs Inc \
\textsuperscript{\rm 4}Microsoft Research\
\textsuperscript{\rm 5}Dalian University of Technology
\\
\{dzliu, xiaoye, panzhou\}@hust.edu.cn, xing.di@protagolabs.com, 
\\ yu.cheng@microsoft.com,  z.xu@dlut.edu.cn
}
\begin{document}
\maketitle
\begin{abstract}
Temporal sentence grounding (TSG) is crucial and fundamental for video understanding. Although the existing methods train well-designed deep networks with a large amount of data, we find that they can easily forget the rarely appeared cases in the training stage due to the off-balance data distribution, which influences the model generalization and leads to undesirable performance. To tackle this issue, we propose a memory-augmented network, called Memory-Guided Semantic Learning Network (MGSL-Net), that learns and memorizes the rarely appeared content in TSG tasks. Specifically, MGSL-Net consists of three main parts: a cross-modal inter-action module, a memory augmentation module, and a heterogeneous attention module. We first align the given video-query pair by a cross-modal graph convolutional network, and then utilize a memory module to record the cross-modal shared semantic features in the domain-specific persistent memory. During training, the memory slots are dynamically associated with both common and rare cases, alleviating the forgetting issue. In testing, the rare cases can thus be enhanced by retrieving the stored memories, resulting in better generalization. At last, the heterogeneous attention module is utilized to integrate the enhanced multi-modal features in both video and query domains. Experimental results on three benchmarks show the superiority of our method on both effectiveness and efficiency, which substantially improves the accuracy not only on the entire dataset but also on rare cases.
\end{abstract}

\section{Introduction}
\begin{figure}[t]
\centering
\includegraphics[width=0.48\textwidth]{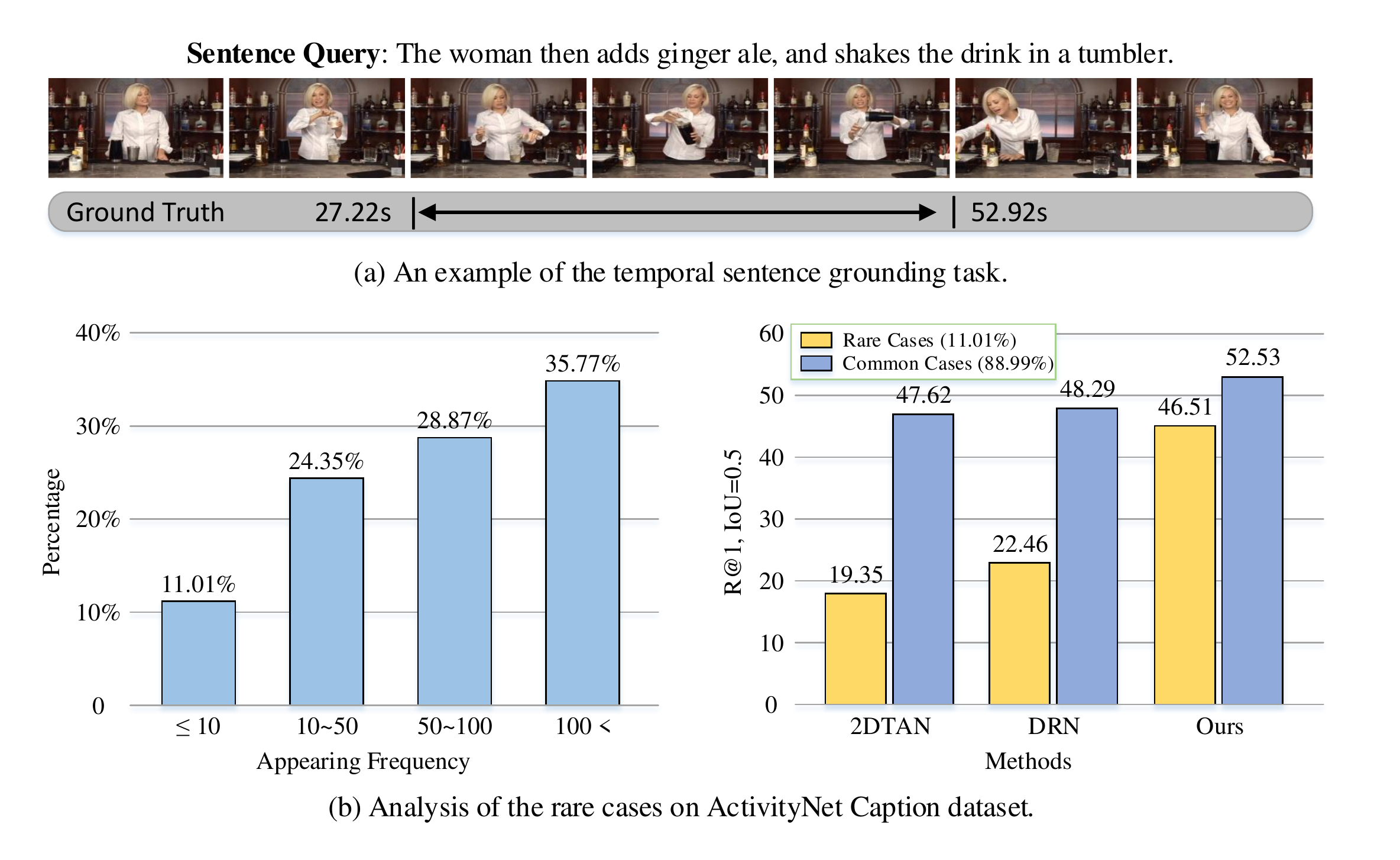}
\caption{(a) An illustrative example of the TSG task. (b) Data distribution on the ActivityNet Caption dataset, and the performance comparison on the corresponding rare cases.}
\label{fig:intro}
\vspace{-10pt}
\end{figure}

Temporal sentence grounding (TSG) is an important yet challenging task in video understanding, which has drawn increasing attention over the last few years due to its vast potential applications in video summarization \cite{song2015tvsum,chu2015video}, video captioning \cite{jiang2018recurrent,chen2020learning}, and temporal action localization \cite{shou2016temporal,zhao2017temporal}, etc. As shown in Figure \ref{fig:intro}, this task aims to ground the most relevant video segment according to a given sentence query. It is substantially more challenging as it needs to not only model the complex multi-modal interactions among video and query features, but also capture complicated context information for their semantics alignment.

Most existing works \cite{anne2017localizing,ge2019mac,liu2018attentive,zhang2019man,chen2018temporally,zhang2019cross,liu2018cross,yuan2019semantic,xu2019multilevel} exploit a proposal-ranking framework that first generates multiple candidate proposals and then ranks them according to their similarities with the sentence query. These methods severely rely on the quality of proposals. Instead of using complex proposals, some recent works \cite{rodriguez2020proposal,yuan2019find,chenrethinking,wang2019language,nan2021interventional,mun2020local,zeng2020dense} utilize a proposal-free framework that directly regresses the temporal locations of the target segment. Compared to the proposal-ranking counterparts, these works are much efficient.

Although the above two types of methods have achieved impressive results, we still can observe their performance bottlenecks on the rarely appeared video-query samples, as shown in Figure \ref{fig:intro}. Here, we select certain pairs of video and sentence as rare samples, which have at least one word (nouns, verbs, or adjectives) whose appearing frequency is less than 10. We can observe that all the existing models can achieve well performance on the common cases, but their performances all drop heavily when evaluated on rare cases. This observation conforms to that deep networks tend to forget the rare cases while learning on a dataset distributed off-balance and diverse \cite{toneva2018empirical}, especially in practical scenarios where the data distribution could be extremely imbalanced. 
To tackle such a challenge, we aim to better match those video-query pair having rarely appeared word-guided semantic for improving the generalization.
However, it is still hard to find a balance between the common and rare samples in the dynamic training process.

To this end, in this paper, we propose to learn and memorize the discriminative and representative cross-modal shared semantic covering all samples, which is implemented by a memory-augmented network, called Memory-Guided Semantic Learning Network (MGSL-Net). Given a pair of video and query input, we first encode their contextual features individually and then align their semantic by a cross-modal graph convolutional network. After obtaining the aligned video-query feature pair, 
we design domain-specific persistent memories in both video and query domains to record cross-modal shared semantic representations which are the most representative. The learned memories are updated and maintained as a compact dictionary shared by all samples. During training, the memory slots in each domain are dynamically associated with both the common and rare samples across mini-batches during the whole training stage, alleviating the forgetting issue. In testing, the rare cases can thus be augmented by retrieving the stored semantic, leading to better generalization. 
Besides, we also develop a heterogeneous attention module to integrate the augmented multi-modal features in video and query domains by considering their contextual inter-modal interactions and video-based self-calibration. 

Our main contributions are summarized as:
\begin{itemize}
    \item We propose a memory-augmented network MGSL-Net for temporal sentence grounding, by learning and memorizing the discriminative and representative cross-modal shared semantics covering all cases. The memory is dynamically associated with both the common and rare samples seen across mini-batches during the whole training, alleviating the forgetting issue on rare samples.
    \item To obtain more domain-specific semantic contexts, we design the memory items in both video and query domains to be persistently read and updated. A heterogeneous attention module is further developed to integrate the enhanced multi-modal features in two domains.
    \item The proposed MGSL-Net achieves state-of-the-art performance on three benchmarks (ActivityNet Caption, TACoS, and Charades-STA), boosting the performance by a large margin not only on the entire dataset but also on the rarely appeared pairwise samples, with limited consumption on computation and memory.
\end{itemize}

\section{Related Work}
\noindent \textbf{Temporal sentence grounding.}
Various algorithms  \cite{anne2017localizing,ge2019mac,liu2018attentive,zhang2019man,chen2018temporally,qu2020fine,liu2021progressively,liu2018cross,liu2021adaptive,liu2022exploring,liu2020jointly,liu2020reasoning} have been proposed within the scan-and-ranking framework, which first generates multiple segment proposals, and then ranks them according to the similarity between proposals and the query to select the best matching one. Some of them \cite{gao2017tall,anne2017localizing} propose
to apply the sliding windows to generate proposals and subsequently integrate the query with segment representations via a matrix operation. 
To improve the quality of the proposals, latest works \cite{wang2019temporally,zhang2019man,yuan2019semantic,zhang2019cross,cao2021pursuit} directly integrate sentence information with each fine-grained video clip unit, and predict the scores of candidate segments by gradually merging the fusion feature sequence over time. 
Instead of generating complex proposals, recent works \cite{rodriguez2020proposal,yuan2019find,chenrethinking,wang2019language,nan2021interventional,mun2020local,zeng2020dense,liu2022unsupervised} directly regress the temporal locations of the target segment. They do not rely on the segment proposals and directly select the starting and ending frames by leveraging cross-modal interactions between video and query. Specifically, they either regress the start/end timestamps based on the entire video representation \cite{yuan2019find,mun2020local}, or predict at each frame to determine whether this frame is a start or end boundary \cite{rodriguez2020proposal,chenrethinking,zeng2020dense}. Although the above methods achieve great performances, they tend to forget the rare cases easily while learning on a dataset distributed off balance and diversely. Different from them, we focus on storing and reading the cross-modal semantic memory to enhance the multi-modal feature representations.

\noindent \textbf{Memory Networks.}
Memory-based approaches have been discussed for solving various problems. NTM \cite{graves2014neural} is proposed to improve the generalization ability of the network by introducing an attention-based memory module. Memory networks like \cite{vaswani2017attention,sukhbaatar2015end} have external memory where information can be further written and read. Xiong \textit{et al.} \cite{xiong2016dynamic} further improve the memory as dynamic memory networks. Different from these unimodal memory models, we propose a cross-modal shared memory which can alternatively interact with multiple data modalities. Although other works \cite{ma2018visual,huang2019acmm} also extend memory networks to multi-modal settings, most of them are
episodic memory networks that are wiped during each forward process. Different from them, our memory model persistently memorizes cross-modal semantic
representations in multi-modal domains with aggregation during the whole training procedure, to better deal with the unbalanced data learning.

\section{Method}
\begin{figure*}[t]
\centering
\includegraphics[width=1.0\textwidth]{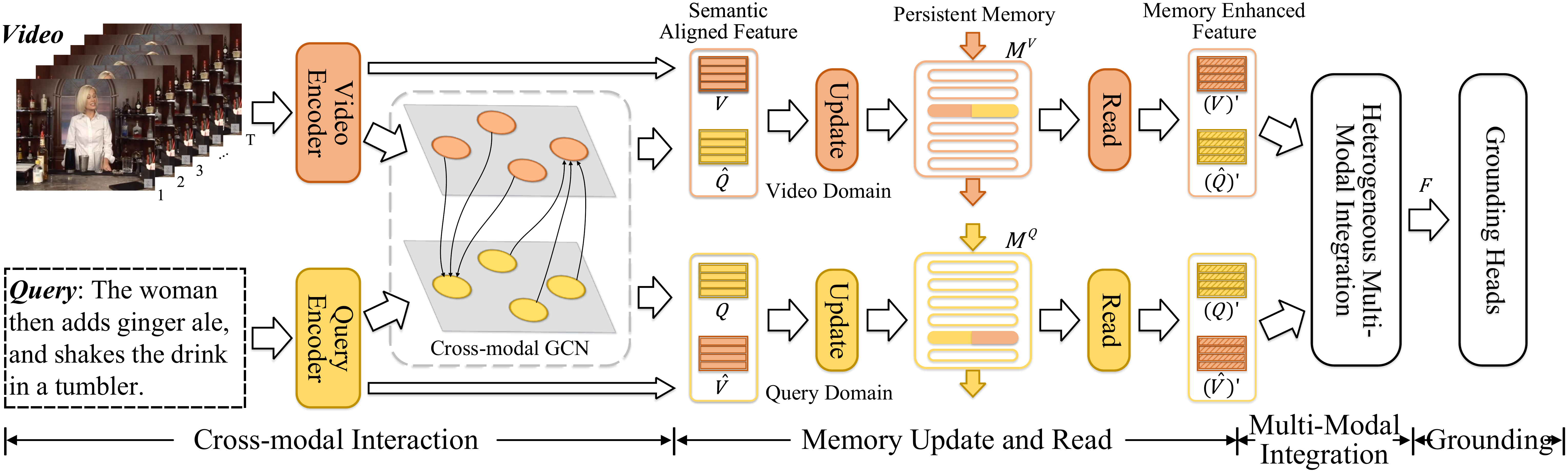}
\caption{Overall pipeline of the proposed MGSL-Net architecture. Given a pair of video and query input, we first encode their features and exploit a cross-modal graph convolutional network (GCN) to align their semantic. Then, for the aligned features in each domain, we utilize a domain-specific persistent memory item to memorize and enhance the cross-modal shared semantic features. After that, we further develop a heterogeneous attention module to integrate multi-modal features in both domains. At last, we locate the target segment by using the regression based grounding heads.}
\label{fig:pipeline}
\vspace{-10pt}
\end{figure*}

\subsection{Overview}
Given an untrimmed video $\mathcal{V}$ and a sentence query $\mathcal{Q}$, we represent the video as $\mathcal{V}=\{v_t\}^{T}_{t=1}$ frame-by-frame, where $v_t$ is the $t$-th frame and $T$ is the number of total frames. Similarly, the query with $N$ words is denoted as $\mathcal{Q}=\{q_n\}^{N}_{n=1}$ word-by-word. The TSG task aims to localize the start and end timestamps $(\tau_s, \tau_e)$ of a specific segment in video $\mathcal{V}$, which refers to the corresponding semantic of query $\mathcal{Q}$.

In this section, we propose a Memory-Guided Semantic Learning Network (MGSL-Net) for TSG task. The overall pipeline of the proposed network, as shown in Figure \ref{fig:pipeline}, includes four main steps:  we first encode both video and query features with contextual information, and align their features with a cross-modal graph convolutional network; then, we utilize the persistent memory items to learn and memorize the cross-modal shared semantic representations in both video and query domains; after getting the memory enhanced multi-modal features, we develop a heterogeneous attention module to consider inter- and intra-modality interactions for multi-modal feature integration; at last, the grounding heads are utilized to localize the segment.

\subsection{Cross-modal Feature Alignment}
\noindent \textbf{Video encoder.}
For video encoding, we first extract the frame-wise features by a pre-trained C3D network \cite{tran2015learning}, and then employ a self-attention \cite{vaswani2017attention} module to capture the long-range dependencies among video frames. We further utilize a BiLSTM \cite{Mike1997} to learn the sequential characteristic. We denote the extracted video features as $\bm{V}=\{\bm{v}_t\}_{t=1}^{T} \in \mathbb{R}^{T \times D}$,where $D$ is the feature dimension.

\noindent \textbf{Query encoder.}
For query encoding, we first generate the word-level features by using the Glove embedding \cite{pennington2014glove}, and also employ a self-attention module and a BiLSTM layer to further encode the query features as $\bm{Q}=\{\bm{q}_n\}_{n=1}^{N} \in \mathbb{R}^{N \times D}$.

\noindent \textbf{Semantic Alignment.}
Considering the obtained video and query representations are intrinsically heterogeneous, we propose a cross-modal graph convolutional network \cite{kipf2016semi} to explicitly perform cross-modal alignment.
Specifically, we first construct two adjacent matrices $\bm{A}_1,\bm{A}_2$ by measuring the cross-modal similarity between each frame-word pair with different directions as:
\begin{equation}
    \bm{A}_1[t,n]= \frac{exp({a_{tn}})}{\sum_{n=1}^N exp({a_{tn}})},
    \bm{A}_2[n,t]= \frac{exp({a_{nt}})}{\sum_{t=1}^T exp({a_{nt}})},
\end{equation}
where value $a_{tn}= a_{nt}^{\top} = \varphi_1(\bm{v}_t) \varphi_2(\bm{q}_n)^{\top}$, $\varphi_1(\cdot),\varphi_2(\cdot)$ are two modality-specific linear mappings to project one modality feature into the same latent space as the other one. $\bm{A}_1 \in \mathbb{R}^{T \times N},\bm{A}_2 \in \mathbb{R}^{N \times T}$ are the normalized adjacent matrices. Therefore, we can get the aligned video representation $\widehat{\bm{V}}$ and the aligned query representation $\widehat{\bm{Q}}$ by:
\begin{equation}
    \widehat{\bm{V}} = \bm{A}_2 \bm{V} \bm{W}_V, \widehat{\bm{Q}} = \bm{A}_1 \bm{Q} \bm{W}_Q, 
\end{equation}
where $\bm{W}_V,\bm{W}_Q \in \mathbb{R}^{D \times D}$ are the weight matrices. $\widehat{\bm{V}}=\{\widehat{\bm{v}}_n\}_{n=1}^{N}$ has the same size $\mathbb{R}^{N \times D}$ as query feature $\bm{Q}$, and they are semantically aligned. For the $n$-th row $\widehat{\bm{v}}_n$ in $\widehat{\bm{V}}$, it is an aggregated representation weighted by cross-modal similarities between the $n$-th word and all the frames. Therefore, $\widehat{\bm{v}}_n$ can be viewed as a visual representation of the $n$-th word, sharing the same semantic meaning with the word $\bm{q}_n$. Similarly, $\widehat{\bm{Q}} = \{\widehat{\bm{q}}_t\}_{t=1}^T \in \mathbb{R}^{T \times D}$ is semantically aligned with $\bm{V}$, and $\widehat{\bm{q}}_t$ can be viewed as a textual representation of the $t$-th frame, sharing the same semantic with the frame $\bm{v}_t$.

\subsection{Memory Network}
Based on the aligned representation pairs ($\widehat{\bm{Q}}$, $\bm{V}$) and ($\widehat{\bm{V}}$, $\bm{Q}$), as shown in Figure \ref{fig:pipeline}, we propose a memory network to learn and memorize the cross-modal shared semantic features in both video domain and query domain, respectively. 

\noindent \textbf{Memory Representation.}
The domain-specific cross-modal shared semantic memories in video and query domains are designed as matrices $\bm{M}^V = \{\bm{m}^V_{l_v}\}_{{l_v}=1}^{L_V} \in \mathbb{R}^{L_V \times D}, \bm{M}^Q =\{\bm{m}^Q_{l_q}\}_{{l_q}=1}^{L_Q} \in \mathbb{R}^{L_Q \times D}$, respectively. Here, $L_V,L_Q$ are the hyper-parameters that defines the number of memory slots and $D$ is the feature dimension. Each memory item $\bm{m}^V_{l_v}$ or $\bm{m}^Q_{l_q}$ can
be updated by intra-domain features with similar semantic meanings, as well as read out to
enhance previously obtained intra-domain features.

Given the aligned frame-word feature pair $(\widehat{\bm{q}}_t,\bm{v}_t)$ from $(\widehat{\bm{Q}},\bm{V})$ in video domain, we aim to interact them with each memory item $\bm{m}^V_{l_v}$ to read and store their shared cross-modal semantic features. Before the interacting process, we first utilize several linear layers to map $\widehat{\bm{q}}_t,\bm{v}_t$ into memory read key, write key, erase value, and write value, respectively. We denote such items of $\widehat{\bm{q}}_t$ as $\bm{k}^{\widehat{Q},r}_t,\bm{k}^{\widehat{Q},w}_t,\bm{e}^{\widehat{Q}}_t,\bm{u}^{\widehat{Q}}_t$, and items of $\bm{v}_t$ as $\bm{k}^{V,r}_t,\bm{k}^{V,w}_t,\bm{e}^{V}_t,\bm{u}^{V}_t$. We also map the aligned features $\widehat{\bm{v}}_n,\bm{q}_n$ into $\bm{k}^{\widehat{V},r}_n,\bm{k}^{\widehat{V},w}_n,\bm{e}^{\widehat{V}}_n,\bm{u}^{\widehat{V}}_n$ and $\bm{k}^{Q,r}_n,\bm{k}^{Q,w}_n,\bm{e}^{Q}_n,\bm{u}^{Q}_n$. Details of how to utilize the generated items to update and read memory will be illustrated as follows.

\noindent \textbf{Updating memory.}
Given the video domain aligned pair $(\widehat{\bm{q}}_t,\bm{v}_t)$ and query domain aligned pair $(\widehat{\bm{v}}_n,\bm{q}_n)$, we determine to write and delete which memory items in $\bm{m}_{l_v}^V$ and $\bm{m}_{l_q}^Q$. 
Specifically, we first calculate the memory addressing weights $w$ according to the similarity between each input feature and corresponding domain-specific memory as:
\begin{equation}
    \footnotesize
    w_{\bm{k}_t,\bm{m}_{l_v}^V} = \frac{exp(s(\bm{k}_t,\bm{m}_{l_v}^V))}{\sum_{l_v} exp(s(\bm{k}_t,\bm{m}_{l_v}^V))}, s(\bm{k}_t,\bm{m}_{l_v}^V) = \frac{\bm{k}_t (\bm{m}_{l_v}^V)^{\top}}{\parallel \bm{k}_t \parallel_2 \parallel\bm{m}_{l_v}^V \parallel_2 },
\end{equation}
\begin{equation}
    \footnotesize
    w_{\bm{k}_n,\bm{m}_{l_q}^Q} = \frac{exp(s(\bm{k}_n,\bm{m}_{l_q}^Q))}{\sum_{l_q} exp(s(\bm{k}_n,\bm{m}_{l_q}^Q))}, s(\bm{k}_n,\bm{m}_{l_q}^Q) = \frac{\bm{k}_n (\bm{m}_{l_q}^Q)^{\top}}{\parallel \bm{k}_n \parallel_2 \parallel\bm{m}_{l_q}^Q \parallel_2 },
\end{equation}
where $\bm{k}_t \in \{\bm{k}^{\widehat{Q},w}_t,\bm{k}^{V,w}_t\}$ and $\bm{k}_n \in \{\bm{k}^{\widehat{V},w}_n,\bm{k}^{Q,w}_n\}$ are the memory write keys, $s(\cdot,\cdot)$ measures the cosine similarity.
Then, we can selectively update memory items by adding new semantic features with write value while deleting old memory with erase value as:
\begin{equation}
    (\bm{m}_{l_v}^V)' = w_{\bm{k}^{V,w}_t,\bm{m}_{l_v}^V} \bm{u}_t^V+\bm{m}_{l_v}^V \odot (1-w_{\bm{k}^{V,w}_t,\bm{m}_{l_v}^V} \bm{e}_t^V),
\end{equation}
\begin{equation}
    (\bm{m}_{l_v}^V)'' = w_{\bm{k}^{\widehat{Q},w}_t,\bm{m}_{l_v}^V} \bm{u}_t^{\widehat{Q}}+(\bm{m}_{l_v}^V)' \odot (1-w_{\bm{k}^{\widehat{Q},w}_t,\bm{m}_{l_v}^V} \bm{e}_t^{\widehat{Q}}),
\end{equation}
\begin{equation}
    (\bm{m}_{l_q}^Q)' = w_{\bm{k}^{Q,w}_n,\bm{m}_{l_q}^Q} \bm{u}_t^Q+\bm{m}_{l_q}^Q \odot (1-w_{\bm{k}^{Q,w}_n,\bm{m}_{l_q}^Q} \bm{e}_t^Q),
\end{equation}
\begin{equation}
    (\bm{m}_{l_q}^Q)'' = w_{\bm{k}^{\widehat{V},w}_n,\bm{m}_{l_q}^Q} \bm{u}_t^{\widehat{V}}+(\bm{m}_{l_q}^Q)' \odot (1-w_{\bm{k}^{\widehat{V},w}_n,\bm{m}_{l_q}^Q} \bm{e}_t^{\widehat{V}}),
\end{equation}
where the erase value $\bm{e}_t \in (0,1)$ is computed with a sigmoid function, and $\odot$ denotes element-wise multiplication. In video domain, $\bm{m}_{l_v}^V$ first updates its memory items with the extracted information from the frame and then from the word. 
In query domain, $\bm{m}_{l_q}^Q$ first updates its memory items with the extracted information from the word and then from the frame. 
In fact, the update order can be alternative and does not show a significant impact on the final performance.

\noindent \textbf{Reading memory.} During the memory reading, we need to read the most relevant items from domain-specific memory item $(\bm{m}_{l_v}^V)''$ and $(\bm{m}_{l_q}^Q)''$ to enhance their representations, respectively. To this end, given the $(\widehat{\bm{q}}_t,\bm{v}_t)$ and $(\widehat{\bm{v}}_n,\bm{q}_n)$, we first compute the cross-modal read weights $w_{\bm{k}^{\widehat{Q},r}_t,(\bm{m}_{l_v}^V)''},w_{\bm{k}^{V,r}_t,(\bm{m}_{l_v}^V)''}$ and $w_{\bm{k}^{\widehat{V},r}_n,(\bm{m}_{l_q}^Q)''},w_{\bm{k}^{Q,r}_n,(\bm{m}_{l_q}^Q)''}$ by comparing read keys with memory items like Eq. (3) and (4). Then we can read memory by regarding the obtained read keys of $(\widehat{\bm{q}}_t,\bm{v}_t)$ and $(\widehat{\bm{v}}_n,\bm{q}_n)$ as queries:
\begin{equation}
    \footnotesize
    (\widehat{\bm{q}}_t)'=\sum_{{l_v}=1}^{L_V} w_{\bm{k}^{\widehat{Q},r}_t,(\bm{m}_{l_v}^V)''} (\bm{m}_{l_v}^V)'', (\bm{v}_t)'=\sum_{{l_v}=1}^{L_V} w_{\bm{k}^{V,r}_t,(\bm{m}_{l_v}^V)''} (\bm{m}_{l_v}^V)'',
\end{equation}
\begin{equation}
    \footnotesize
    (\widehat{\bm{v}}_n)'=\sum_{{l_q}=1}^{L_Q} w_{\bm{k}^{\widehat{V},r}_n,(\bm{m}_{l_q}^Q)''} (\bm{m}_{l_q}^Q)'', (\bm{q}_n)'=\sum_{{l_q}=1}^{L_Q} w_{\bm{k}^{Q,r}_n,(\bm{m}_{l_q}^Q)''} (\bm{m}_{l_q}^Q)'',
\end{equation}
where $(\widehat{\bm{q}}_t)',(\bm{v}_t)'$ and $(\widehat{\bm{v}}_n)',(\bm{q}_n)'$ are the read
vectors which can be regarded as memory-enhanced representations of video and query domains, respectively.

\begin{figure}[t]
\centering
\includegraphics[width=0.48\textwidth]{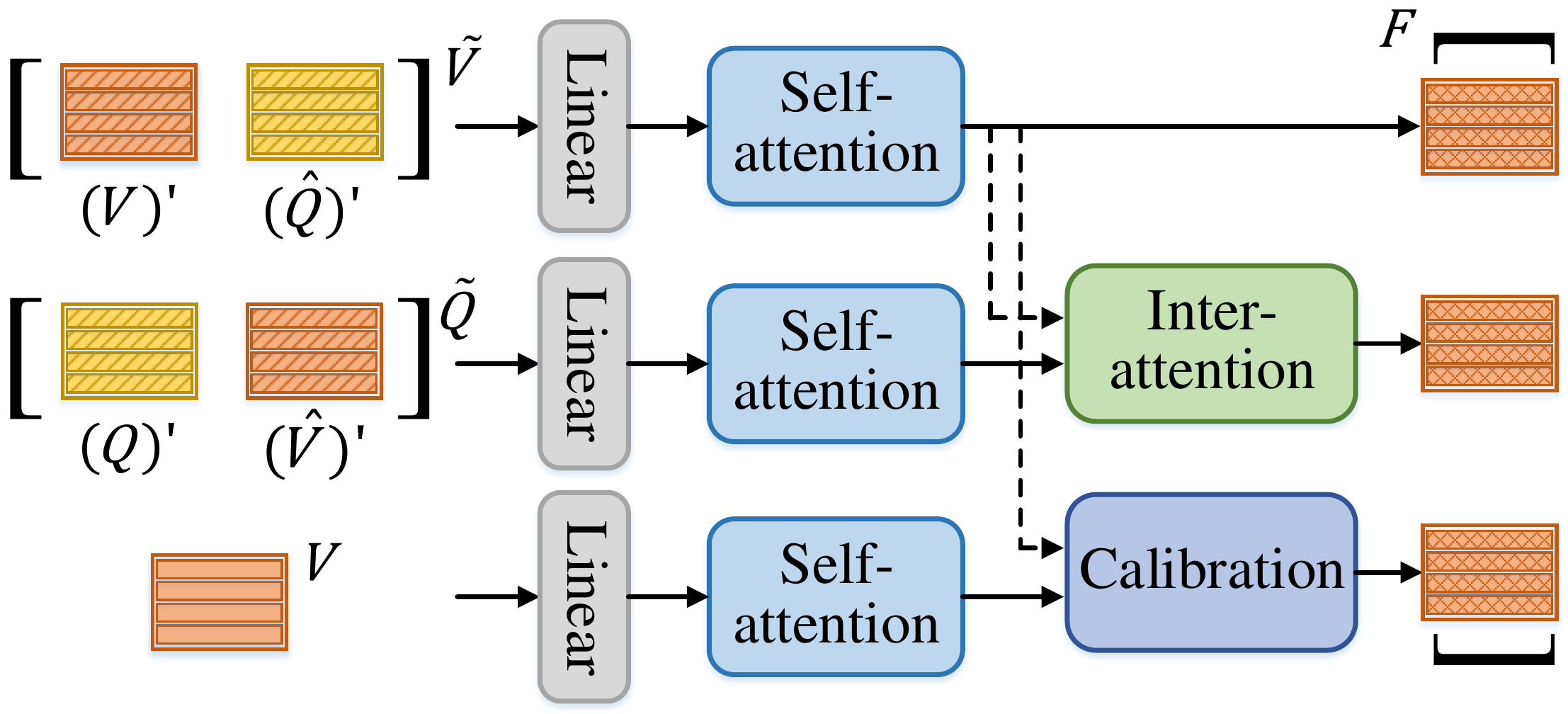}
\caption{Illustration of Heterogeneous Attention module. The units of self-attention, inter-attention and calibration are implemented by dot product attention.}
\label{fig:integration}
\vspace{-10pt}
\end{figure}

\subsection{Heterogeneous Multi-Modal Integration}
After obtaining memory enhanced representations from both video and query domains, we generate new video representation $\widetilde{\bm{V}}=\{\widetilde{\bm{v}}_t\}_{t=1}^T \in \mathbb{R}^{T \times 2D}$ and new query representation $\widetilde{\bm{Q}}=\{\widetilde{\bm{q}}_n\}_{n=1}^N \in \mathbb{R}^{N \times 2D}$ by concatenation operation, where $\widetilde{\bm{v}}_t=[(\bm{v}_t)';(\widehat{\bm{q}}_t)']$ and
$\widetilde{\bm{q}}_n=[(\bm{q}_n)';(\widehat{\bm{v}}_n)']$. To further integrate these two representations, we develop a heterogeneous attention module to consider their inter- and intra-modality interactions.
In particular, we additionally take the original video feature $\bm{V}$ as global context to calibrate the learned memory contents in $\widetilde{\bm{V}}$. As shown in Figure \ref{fig:integration}, the proposed heterogeneous attention mechanism first utilizes three linear layers to map the three representations in to the same latent space, and then exploits a self-attention unit to capture the semantic-aware intra-modality relations between the enhanced frame-frame pairs and word-word pairs. After that, the inter-modality relationship is captured by interacting the features of frame-word pair. To further calibrate the memory-wise contents, we take $\bm{V}$ as the global signals to supervise the enhanced video feature $\widetilde{\bm{V}}$. These three attentional units are combined in a modular way in defining the heterogeneous attention mechanism, and all the units are based on the dot product attention. Finally, the integrated feature $\bm{F}=\{\bm{f}_t\}_{t=1}^T$ is obtained by concatenating all those output features.

\subsection{Grounding Heads}
Taking the fine-grained feature $\bm{F}=\{\bm{f}_t\}_{t=1}^T$ as input, we process frame-wise feature $\bm{f}_t$ by the grounding module, which consists of three components: boundary regression head, confidence scoring head and IoU regression head. Since TSG task aims to localize a specific segment, the boundary regression head is designed to predict the temporal bounding box at each frame. To select the box that matches the query best, we propose the confidence scoring head to predict scores indicating whether the content in each bbox matches the query semantically. A IoU regression head is also utilized to predict score for directly estimating the IoU between each bbox and the ground truth segment.

\noindent \textbf{Boundary regression head.} We implement this head as two 1D convolution layers with two output channels, and we only assign regression targets for positive frames. If location $t$ falls inside the ground truth $(\tau_s,\tau_e)$, the regression targets are $d_t=(d_{t,s},d_{t,e})$, where $d_{t,s}=t-\tau_s,d_{t,e}=\tau_e-t$. For the predicted $\hat{d}_t$ and ground truth $d_t$, we define $\mathcal{L}_b$ as:
\begin{equation}
    \footnotesize
    \mathcal{L}_b = \frac{1}{T_p}\sum_{t=1}^T \mathds{1}_t (\mathcal{L}_1(d_t,\hat{d}_t)-ln \frac{{min}(d_{t,e},\hat{d}_{t,e})-{max}(d_{t,s},\hat{d}_{t,s})}{{max}(d_{t,e},\hat{d}_{t,e})-{min}(d_{t,s},\hat{d}_{t,s})}),
\end{equation}
where $\mathcal{L}_1$ is a smooth $l_1$ loss, the second item is a IoU loss. $\mathds{1}_t$ is the indicator function, being 1 if frame $t$ is positive and 0 otherwise. $T_p$ is the number of  positive frames.

\noindent \textbf{Confidence scoring head.} This head is implemented as two 1D convolution layers with one output channel. For each frame t, if it falls in the ground truth, we think its generated bbox matches the query semantically and denote its label as $c_t=1$. If not, we denote it as $c_t=0$. We utilize a a binary cross entropy loss for confidence evaluation as:
\begin{equation}
    \mathcal{L}_c = \frac{1}{T_p} \sum_{t=1}^T \mathcal{L}_{bce}(c_t,\hat{c}_t).
\end{equation}

\noindent \textbf{IoU regression head.} We train a three-layer 1D convolution to estimate the IoU between the generated bbox at each frame and the corresponding ground truth. Denoting the ground truth IoU as $i_t$ and predicted one as $\hat{i}_t$, we have:
\begin{equation}
    \mathcal{L}_i = \frac{1}{T} \sum_{t=1}^T \mathcal{L}_1 (i_t,\hat{i}_t).
\end{equation}

Thus, the final loss is a multi-task loss combing the above three loss functions as:
\begin{equation}
    \mathcal{L} = \lambda_1 \mathcal{L}_b + \lambda_2 \mathcal{L}_c + \lambda_3 \mathcal{L}_i,
\end{equation}
where $\lambda_{1}, \lambda_{2}$ and $\lambda_{3}$ are the hyper-paremeters to balance the training weights on different losses.

\section{Experiments}
\subsection{Datasets and Evaluation}
\noindent \textbf{ActivityNet Caption.}
ActivityNet Caption \cite{krishna2017dense} contains 20000 untrimmed videos with 100000 descriptions from YouTube. The videos are 2 minutes on average, and the annotated video clips have much larger variation, ranging from several seconds to over 3 minutes. Following public split, we use 37,417, 17,505, and 17,031 sentence-video pairs for training, validation, and testing respectively.

\noindent \textbf{TACoS.}
TACoS \cite{regneri2013grounding} is widely used on TSG task and contain 127 videos. The videos from TACoS are collected from cooking scenarios, thus lacking the diversity. They are around 7 minutes on average. We use the same split as \cite{gao2017tall}, which includes 10146, 4589, 4083 query-segment pairs for training, validation and testing.

\noindent \textbf{Charades-STA.}
Charades-STA is built on the Charades dataset \cite{sigurdsson2016hollywood}, which focuses on indoor activities. In total, there are 12408 and 3720 moment-query pairs in the training and testing sets respectively.

\noindent \textbf{Evaluation Metric.}
Following previous works \cite{gao2017tall,zeng2020dense,zhang2019learning}, we adopt “R@n, IoU=m” as our evaluation metrics. It is defined as the percentage of at least one of top-n selected moments having IoU larger than m, which is the higher the better.

\subsection{Implementation Details}
We utilize the $112 \times 112$ pixels shape of every frame of videos as input, and apply C3D \cite{tran2015learning} to encode the videos on ActivityNet Caption, TACoS, and I3D \cite{carreira2017quo} on Charades-STA. 
We set the length of video feature sequences to 200 for ActivityNet Caption and TACoS datasets, 64 for Charades-STA dataset. As for sentence encoding, we set the length of word feature sequences to 20, and utilize Glove embedding \cite{pennington2014glove} to embed each word to 300 dimension features. The hidden state dimension of BiLSTM networks is set to 512. The number of memory items $(L_V,L_Q)$ are set to (1024,1024), (512,512), (512,512) for three datasets, respectively. We empirically find that further increasing the memory number results in a convergence of the performance. The balanced weights of $\mathcal{L}$ are $\lambda_1 = \lambda_2 = \lambda_3 = 1.0$. During the training, we use an Adam optimizer with the leaning rate of 0.0001. The model is trained for 50 epochs to guarantee its convergence with a batch size of 128. All the experiments are implemented on a single NVIDIA TITAN XP GPU.

\begin{table*}[t!]
    \small
    \centering
    \setlength{\tabcolsep}{1.2mm}{
    \begin{tabular}{c|cccc|cccc|cccc}
    \hline \hline
    \multirow{3}*{Method} & \multicolumn{4}{c|}{ActivityNet Captions} & \multicolumn{4}{c|}{TACoS} & \multicolumn{4}{c}{Charades-STA} \\ \cline{2-5} \cline{6-9} \cline{10-13}
    ~ & R@1, & R@1, & R@5, & R@5, & R@1, & R@1, & R@5, & R@5, & R@1, & R@1, & R@5, & R@5, \\ 
    ~ & IoU=0.5 & IoU=0.7 & IoU=0.5 & IoU=0.7 & IoU=0.3 & IoU=0.5 & IoU=0.3 & IoU=0.5 & IoU=0.5 & IoU=0.7 & IoU=0.5 & IoU=0.7 \\ \hline
    TGN & 28.47 & - & 43.33 & - & 21.77 & 18.90 & 39.06 & 31.02 & - & - & - & -\\
    CTRL & 29.01 & 10.34 & 59.17 & 37.54 & 18.32 & 13.30 & 36.69 & 25.42 & 23.63 & 8.89 & 58.92 & 29.57 \\
    ACRN & 31.67 & 11.25 & 60.34 & 38.57  & 19.52 & 14.62 & 34.97 & 24.88 & 20.26 & 7.64 & 71.99 & 27.79 \\
    QSPN & 33.26 & 13.43 & 62.39 & 40.78 & 20.15 & 15.23 & 36.72 & 25.30 & 35.60 & 15.80 & 79.40 & 45.40 \\
    CBP & 35.76 & 17.80 & 65.89 & 46.20 & 27.31 & 24.79 & 43.64 & 37.40 & 36.80 & 18.87 & 70.94 & 50.19 \\
    SCDM & 36.75 & 19.86 & 64.99 & 41.53 & 26.11 & 21.17 & 40.16 & 32.18 & 54.44 & 33.43 & 74.43 & 58.08 \\
    GDP & 39.27 & - & - & - & 24.14 & - & - & - & 39.47 & 18.49 & - & - \\
    LGI & 41.51 & 23.07 & - & - & - & - & - & - & 59.46 & 35.48 & - & - \\
    BPNet & 42.07 & 24.69 & - & - & 25.96 & 20.96 & - & - & 50.75 & 31.64 & - & - \\
    VSLNet & 43.22 & 26.16 & - & - & 29.61 & 24.27 & - & - & 54.19 & 35.22 & - & - \\
    CMIN & 43.40 & 23.88 & 67.95 & 50.73 & 24.64 & 18.05 & 38.46 & 27.02 & - & - & - & - \\
    2DTAN & 44.51 & 26.54 & 77.13 & 61.96 & 37.29 & 25.32 & 57.81 & 45.04 & 39.81 & 23.25 & 79.33 & 51.15 \\
    DRN & 45.45 & 24.36 & 77.97 & 50.30 & - & 23.17 & - & 33.36 & 53.09 & 31.75 & 89.06 & 60.05 \\
    CBLN & 48.12 & 27.60 & 79.32 & 63.41 & 38.98 & 27.65 & 59.96 & 46.24 & 61.13 & 38.22 & 90.33 & 61.69 \\ \hline
    \textbf{MGSL-Net} & \textbf{51.87} & \textbf{31.42} & \textbf{82.60} & \textbf{66.71} & \textbf{42.54} & \textbf{32.27} & \textbf{63.39} & \textbf{50.13} & \textbf{63.98} & \textbf{41.03} & \textbf{93.21} & \textbf{63.85} \\ \hline \hline
    \end{tabular}}
    \caption{Performance compared with the state-of-the-arts on ActivityNet Caption, TACoS, and Charades-STA datasets.}
    \vspace{-10pt}
    \label{tab:compare}
\end{table*}

\subsection{Comparisons with the State-of-the-Arts}
\noindent \textbf{Compared Methods.}
We compare the proposed MGSL-Net with
state-of-the-art TSG methods on three datasets. These
methods are grouped into three categories by the viewpoints
of proposal-based and proposal-free approach: 1) proposal-based methods: TGN \cite{chen2018temporally}, CTRL \cite{gao2017tall}, ACRN \cite{liu2018attentive}, QSPN \cite{xu2019multilevel}, CBP \cite{wang2019temporally}, SCDM \cite{yuan2019semantic}, CMIN \cite{zhang2019cross}, 2DTAN \cite{zhang2019learning}, and CBLN \cite{liu2021context}. 2) proposal-free methods: GDP \cite{chenrethinking}, LGI \cite{mun2020local}, VSLNet \cite{zhang2020span}, DRN \cite{zeng2020dense}. 3) others: BPNet \cite{xiao2021boundary}. Note that all the above methods directly utilize deep networks to learn cross-modal retrieval without considering the rarely appeared  video-query samples.

\noindent \textbf{Comparison on ActivityNet Caption.}
Table \ref{tab:compare} summarizes the results on ActivityNet Caption. It shows that our MGSL-Net outperforms all the baselines in all metrics. 
Specifically, we observe that MGSL-Net works well in even
stricter metrics, e.g., it achieved a significant 3.82\% and 3.30\% absolute improvement in R@1, and R@5, IoU=0.7 compared to the previous state-of-the-art method CBLN, which demonstrates the superiority of our model.
It is mainly because our memory can store useful cross-modal
shared semantic representations, and thus better associate
those rarely appeared video and query in the standard
test sets.

\noindent \textbf{Comparison on TACoS.}
From Table \ref{tab:compare}, we can also find that our MGSL-Net achieves the best performance on TACoS dataset. Note that our model shows much larger improvements on the TACoS dataset than the ActivityNet Caption dataset. It mainly results from that the fewer training data with low diversity of TACoS dataset cannot guarantee the previous models can well capture the relations among small object. But our model can better exploit the
auxiliary resources for better learning.

\noindent \textbf{Comparison on Charades-STA.} As shown in Table \ref{tab:compare}, our MGSL-Net achieves new state-of-the-art performance over all metrics on Charades-STA. Since there exists less rarely appeared samples in this dataset, it has less performance improvements than the other two datasets.

\begin{table}[t!]
    \small
    \centering
    \setlength{\tabcolsep}{1.5mm}{
    \begin{tabular}{cccc}
    \hline \hline
    Method & Run-Time & Model Size & R@1, IoU=0.5 \\ \hline
    ACRN & 4.31s & 128M & 14.62 \\
    CTRL & 2.23s & \textbf{22M} & 13.30 \\ 
    TGN & 0.92s & 166M & 18.90 \\
    2DTAN & 0.57s & 232M & 25.32\\ 
    DRN & 0.15s & 214M & 23.17 \\ \hline
    \textbf{MGSL-Net} & \textbf{0.10s} & 203M & \textbf{32.27} \\ \hline
    \end{tabular}}
    \caption{Efficiency comparison run on TACoS dataset.}
    \vspace{-10pt}
    \label{tab:efficient}
\end{table}

\subsection{Efficiency Comparison}
We evaluate the efficiency of our MGSL-Net, by fairly comparing its running time and parameter size with existing methods on a single Nvidia TITAN XP GPU on TACoS dataset. As shown in Table \ref{tab:efficient}, it can be observed that we achieves much faster processing speeds and relatively less learnable parameters. This attributes to: 1) The proposal generation procedure and proposal matching procedure of proposal-based methods (ACRN, CTRL, TGN, 2DTAN) are quite time-consuming. 2) Regression-based method DRN utilizes much convolutional layers to achieve multi-level feature fusion for cross-modal interaction, which is also cost time. 3) Our MGSL is free from complex and time-consuming operations, showing superiority in both effectiveness and efficiency.

\begin{figure}[t]
\centering
\includegraphics[width=0.48\textwidth]{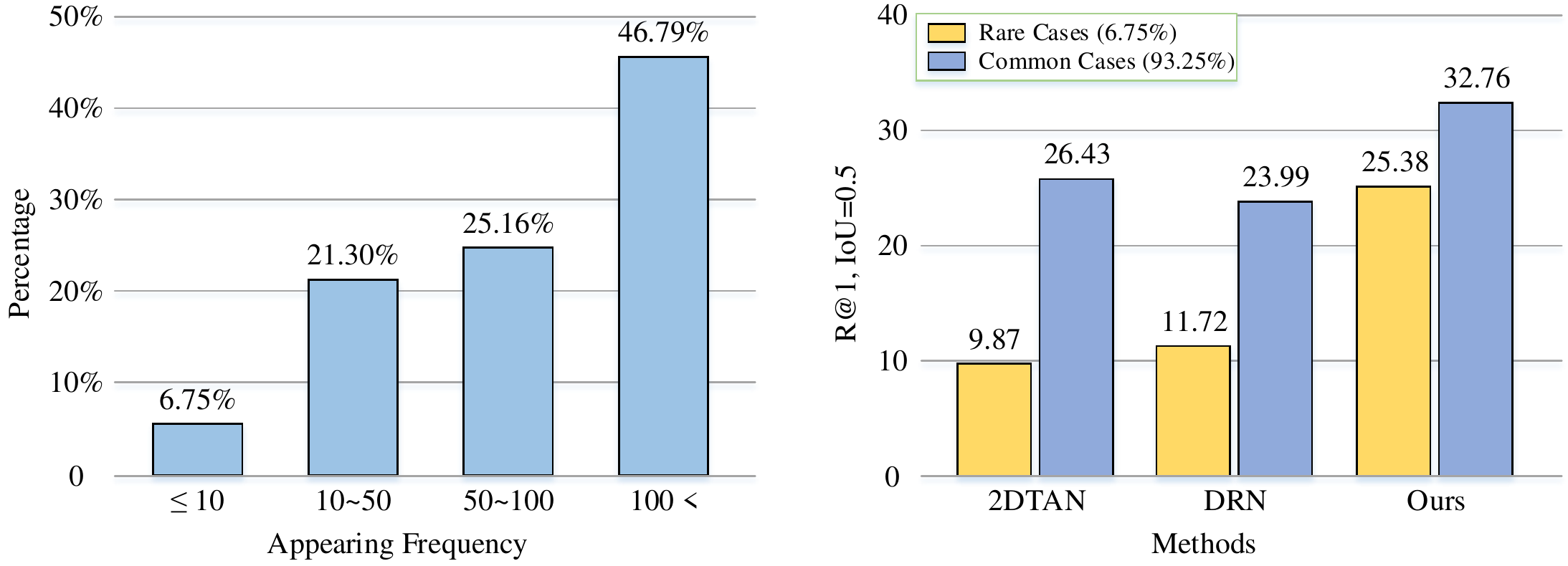}
\caption{Data distribution on the TACoS dataset, and the performance comparison on its rare cases.}
\label{fig:tacos}
\vspace{-6pt}
\end{figure}

\begin{figure}[t]
\centering
\includegraphics[width=0.48\textwidth]{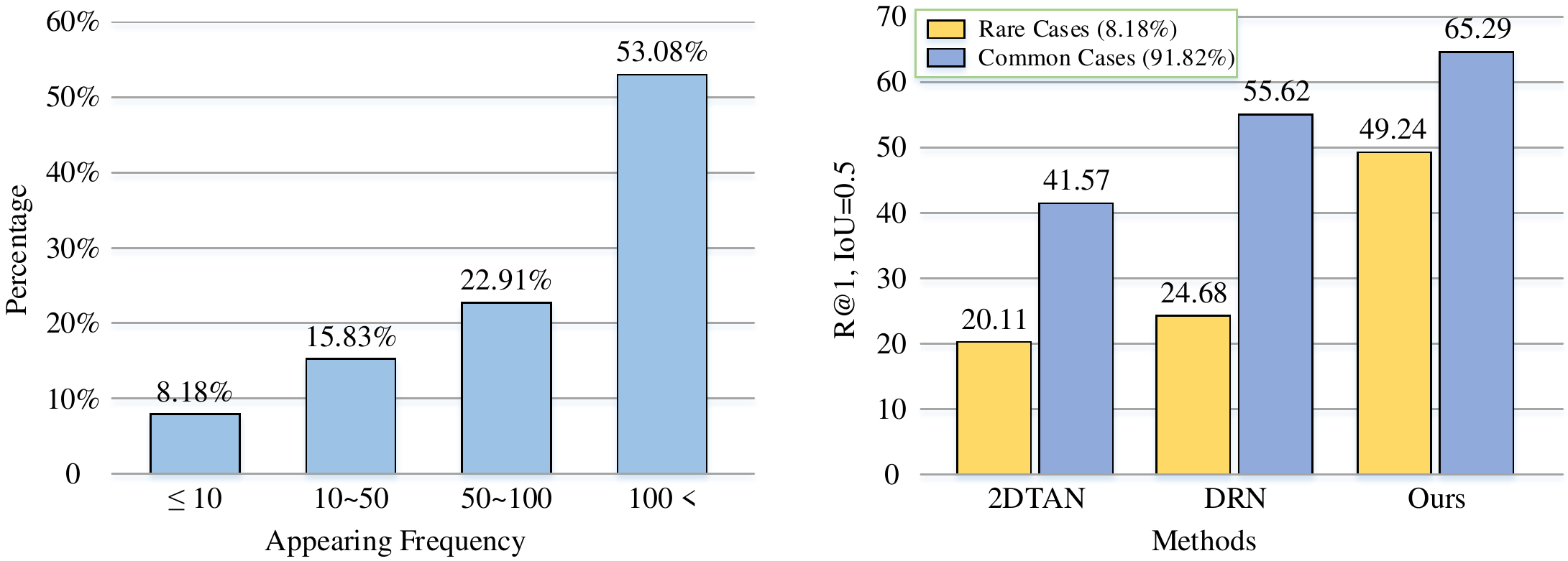}
\caption{Data distribution on the Charades-STA dataset, and the performance comparison on its rare cases.}
\label{fig:charades}
\vspace{-10pt}
\end{figure}

\subsection{Analysis on the Rare Cases}
\begin{figure}[t]
\centering
\includegraphics[width=0.48\textwidth]{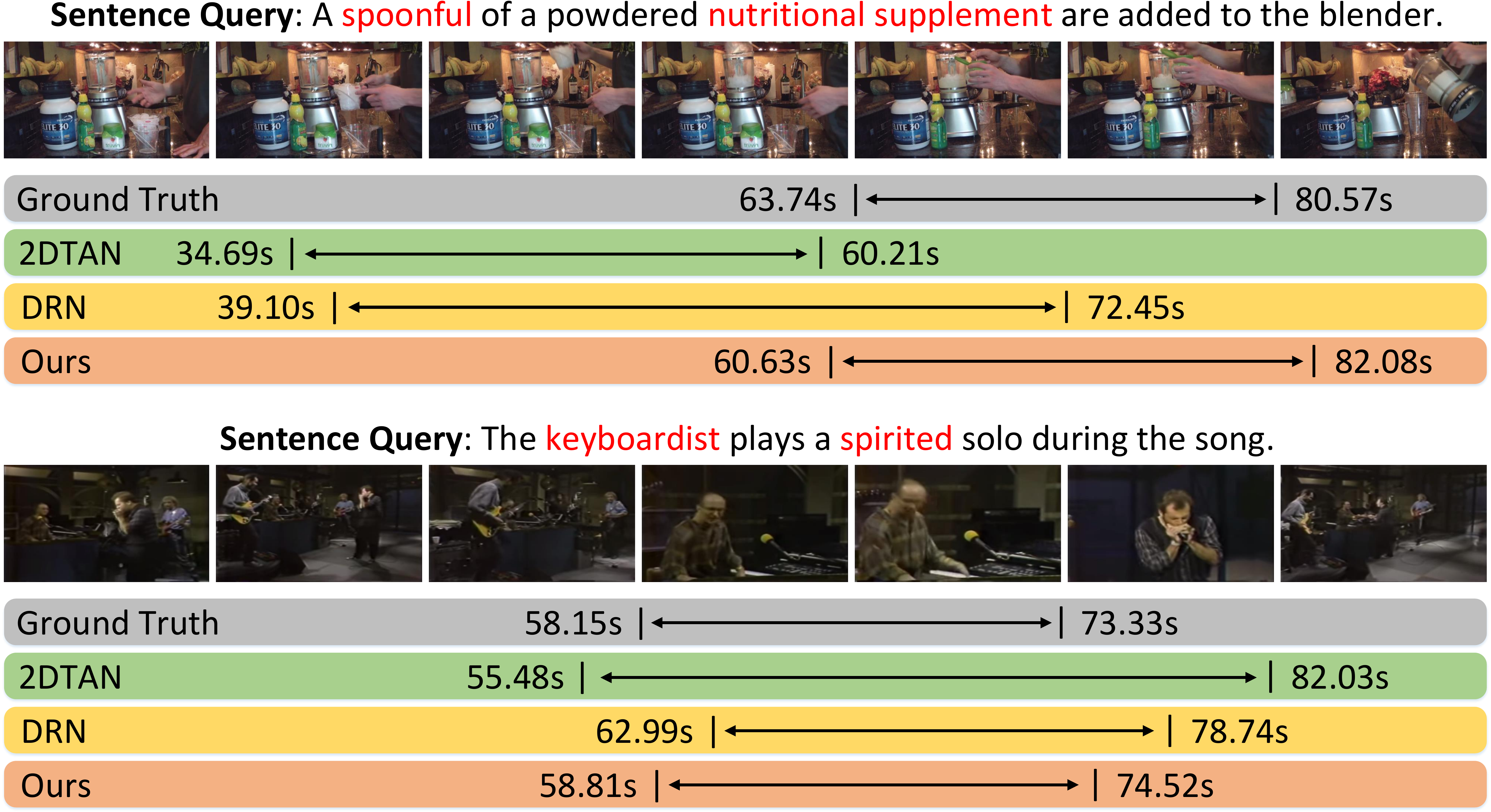}
\caption{The qualitative results of our proposed method. Rarely appeared words are marked as red.}
\label{fig:result}
\end{figure}

Analyzing with the rarely appeared video-query samples is not easy since such pair-wise data is not easy to define. 
Therefore, we first analyze the data distribution of each dataset as shown in Figure \ref{fig:intro}, \ref{fig:tacos} and \ref{fig:charades}, and then we select certain pairs of video and sentence as rare samples, which have at least one word (nouns, verbs or adjectives) whose appearing frequency is less than 10. The other remained samples are treated as common samples.
In these three figures, we show the performance of different methods on the rare cases and common cases. Our proposed method is more effective to handle the rare cases, which brings much more improvements than other methods. We also give the qualitative examples of the grounding results as shown in Figure \ref{fig:result}, where our method can learn and memorize the semantics of the rare cases and can ground the segment more accurately.

\begin{table}[t!]
    \small
    \centering
    \setlength{\tabcolsep}{1.5mm}{
    \begin{tabular}{cc|cc}
    \hline \hline
    Domain-specific & Heterogeneous & R@1, & R@1, \\ 
    memory networks & attention module & IoU=0.5 & IoU=0.7 \\ \hline
      & & 43.65 & 22.48 \\
    $\checkmark$ & & 49.24 & 28.01 \\
      & $\checkmark$ & 46.19 & 25.73 \\
    \hline
     $\checkmark$ & $\checkmark$ & \textbf{51.87} & \textbf{31.42} \\ \hline \hline
    \end{tabular}}
    \caption{Main ablation study on ActivityNet Caption dataset.}
    \label{tab:ablation1}
\end{table}

\begin{table}[t!]
    \small
    \centering
    \setlength{\tabcolsep}{1.5mm}{
    \begin{tabular}{ccc|ccc|cc}
    \hline \hline
    \multicolumn{3}{c|}{Video domain} & \multicolumn{3}{c|}{Query domain} & R@1, & R@1, \\ \cline{1-6}
    $\bm{V}$ & $\widehat{\bm{Q}}$ & shared & $\bm{Q}$ & $\widehat{\bm{V}}$ & shared & IoU=0.5 & IoU=0.7 \\ \hline
    & & & & & & 46.19 & 25.73 \\
    $\checkmark$ & & & & & & 47.64 & 26.82 \\
    $\checkmark$ & $\checkmark$ & & & & & 48.88 & 27.90 \\
    $\checkmark$ & $\checkmark$ & $\checkmark$ & & & & 49.97 & 28.21 \\
    $\checkmark$ & $\checkmark$ & $\checkmark$& $\checkmark$ & & & 50.56 & 29.60 \\
    $\checkmark$ & $\checkmark$ & $\checkmark$& $\checkmark$ & $\checkmark$ & & 51.20 & 30.85 \\
    \hline
    $\checkmark$ & $\checkmark$ &$\checkmark$  & $\checkmark$ & $\checkmark$ & $\checkmark$ & \textbf{51.87} & \textbf{31.42} \\ \hline \hline
    \end{tabular}}
    \caption{Performance comparisons with the memory network in different settings on ActivityNet Caption dataset.}
    \label{tab:ablation2}
\end{table}

\begin{table}[t!]
    \small
    \centering
    \setlength{\tabcolsep}{1.2mm}{
    \begin{tabular}{ccc|cc}
    \hline \hline
    Self- & Inter- & \multirow{2}*{Calibration} & R@1, & R@1, \\ 
    attention & attention & ~ & IoU=0.5 & IoU=0.7 \\ \hline
     & $\checkmark$ &  &  49.24 & 28.01 \\ 
    & $\checkmark$ & $\checkmark$ & 50.33 & 28.98 \\
    $\checkmark$ & $\checkmark$ & & 51.27 & 30.56 \\
    \hline
    $\checkmark$ & $\checkmark$ & $\checkmark$ & \textbf{51.87} & \textbf{31.42} \\ \hline \hline
    \end{tabular}}
    \caption{Performance comparisons with the heterogeneous attention module in different settings on ActivityNet Caption dataset.}
    \vspace{-10pt}
    \label{tab:ablation3}
\end{table}

\subsection{Ablation Study}
\noindent \textbf{Main ablation.} As shown in Table \ref{tab:ablation1}, we first study the influence of each main component in our proposed MGSL-Net. We set the MGSL-Net model without both domain-specific memory networks and heterogeneous multi-modal integration module as the baseline. The table shows that both memory network and heterogeneous attention make great contributions for the final performance, where the memory network brings the largest improvement of 5-6 absolute values.

\noindent \textbf{Investigation on memory network.} We investigate the performance comparison with the memory network with different settings as shown in Table \ref{tab:ablation2}, where ``shared" means utilizing a shared memory slots to learn the semantics of two inputs in each domain. From the table, we can find several points: 1) The two aligned semantics in pairwise data $(\bm{V},\widehat{\bm{Q}})$ or $(\bm{Q},\widehat{\bm{V}})$ are all important for memory learning. 2) In each domain, the shared memory performs better than utilizing two separated memories for reading and updating the pairwise data. 3) The memory-based contexts in both video and query domains are all helpful for grounding.

\begin{figure}[t]
\centering
\includegraphics[width=0.48\textwidth]{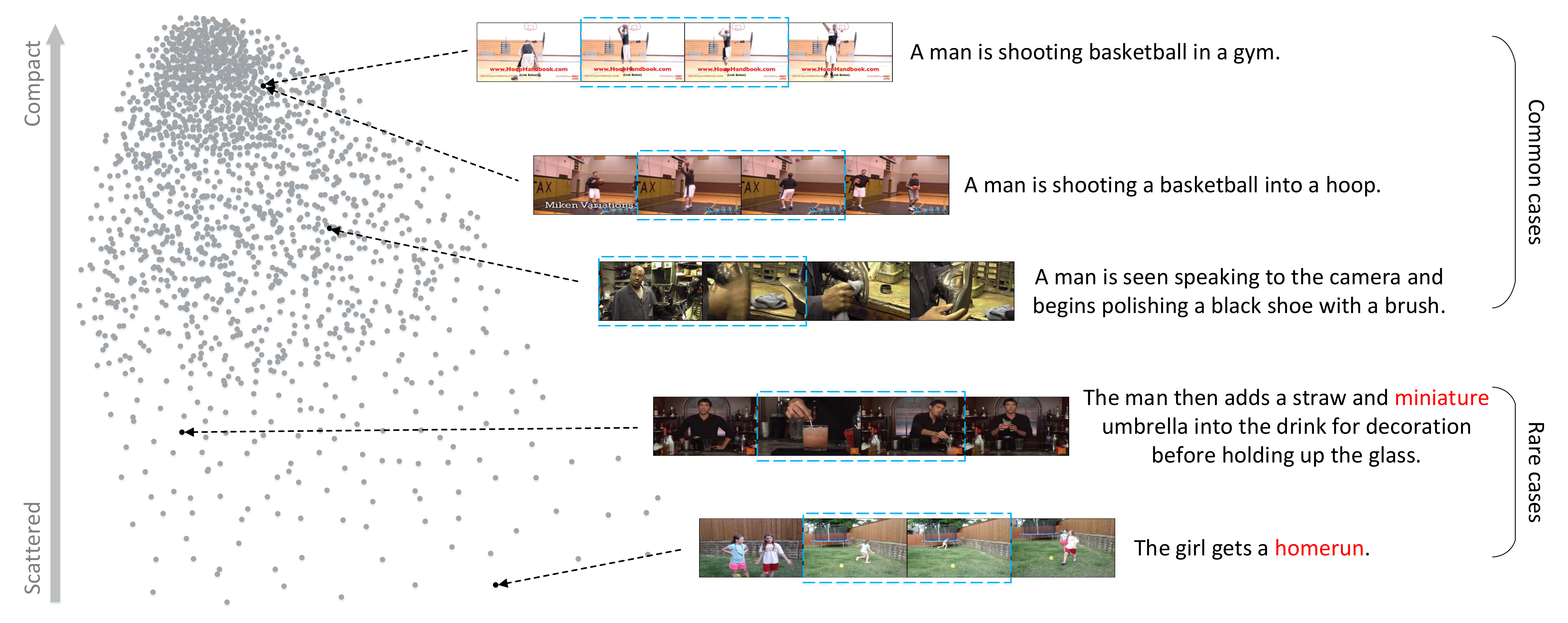}
\caption{Two-dimensional visualization of learned memory items. The blue rectangle denotes the target segment. Rarely appeared words are marked as red.}
\label{fig:memory}
\vspace{-10pt}
\end{figure}

To further investigate what the shared memory actually learns, we reduce the dimensionality of memory slots with PCA, and show their two-dimensional representations (grey nodes) in Figure \ref{fig:memory}. We can see that all the nodes distribute in a divergent shape, in which the top nodes are more compact while the bottom ones are more scattered. To figure out the semantic meanings of these memory slots, we take several representative nodes (with arrows) as queries to retrieve video-query pair. 
We find the rarely appeared content is indeed captured and represented as the scattered nodes while more commonly appeared content is captured and represented as the compact ones.

\noindent \textbf{Investigation on heterogeneous attention.}
We also conduct the ablation studies on the heterogeneous attention module in Table \ref{tab:ablation3}, where we set the inter-attention branch as the baseline. The self-attention brings largest improvement, since it not only captures the intra-relations among the elements in each modality but also provides the enhanced video features in video domain. The calibration module also makes contribution to the final performance.

\section{Conclusion}
In this paper, we have proposed the Memory-Guided Semantic Learning (MGSL) to handle the rarely appeared pairwise samples in temporal sentence grounding task. The main contributions of
this work are: 1) we propose a cross-modal graph convolutional network to align the semantic between video and query, 2) we develop two domain-specific persistent memory items to learn and memorize the cross-modal shared semantic representations, and 3) we devise a heterogeneous attention module to integrate the enhanced multi-modal features in both video and query domains. Experimental results shows the superiority of our method on both effectiveness and efficiency.

\section{Acknowledgements} 
This work was supported in part by the National Natural Science Foundation of China (NSFC) under Grant (No.61972448, No.62172068, No.61802048).

\bibliography{reference.bib}

\end{document}